%% file: main.tex
\documentclass[runningheads]{llncs}

\usepackage{eccvabbrv}

\usepackage{graphicx}
\usepackage{booktabs}

\usepackage{caption}
\usepackage[accsupp]{axessibility}  

\usepackage[pagebackref,breaklinks,colorlinks,citecolor=eccvblue]{hyperref}
\usepackage{hypcap} 

\usepackage{orcidlink}
\usepackage{times}
\usepackage{epsfig}
\usepackage{graphicx}
\usepackage{amsmath}
\usepackage{amsmath, amssymb}

\usepackage{amssymb}
\usepackage{float}
\usepackage[numbers,sort,compress]{natbib}
\usepackage{graphicx,booktabs}
\usepackage{xcolor}
\usepackage{amsmath}
\usepackage{relsize}
\usepackage{xcolor}
\usepackage{graphicx}
\usepackage{appendix}

\usepackage[font=scriptsize]{caption}

\usepackage{hyperref}
\hypersetup{colorlinks,linkcolor={red},citecolor={green},urlcolor={red}} 

\usepackage{amsfonts}
\usepackage{subcaption}
\usepackage{pifont}
\usepackage{amsmath,  amssymb}
\usepackage{float}

\newcommand{\n}[1]{\textcolor{red}{#1}}
\newcommand{\bl}[1]{\textcolor{blue}{#1}}
\newcommand{\orange}[1]{\textcolor{orange}{#1}}

\renewcommand{\vec}[1]{\ensuremath{\mathbf{#1}}}
\newcommand{\mat}[1]{\ensuremath{\mathbf{#1}}}

\newcommand{\transpose}{\ensuremath{^\top}}

\newcommand{\bigsum}[3]{\mathlarger{\sum}_{#1 = #2}^{#3}}

\usepackage{xcolor}

\definecolor{aogreen}{rgb}{0.0, 0.5, 0.0}

\newcommand{\mob}{\ensuremath{\operatorname{M\ddot{o}bius}}}

\begin{document}

\pagestyle{headings}
\mainmatter
\def\ECCV16SubNumber{2132}  

\title{Occlusion Handling in 3D Human Pose Estimation with Perturbed Positional Encoding} 

\titlerunning{PerturbPE}

\authorrunning{Azizi et al.}

\author{Niloofar Azizi\inst{1} \and
Mohsen Fayyaz\inst{2} \and
Horst Bischof\inst{1}}
\institute{Graz University of Technology, Graz, Austria
\email{\{azizi, bischof\}@tugraz.at}\\
 \and
 Microsoft\\
\email{mohsenfayyaz@microsoft.com}}

\maketitle

\begin{abstract}
Understanding human behavior fundamentally relies on accurate 3D human pose estimation. Graph Convolutional Networks (GCNs) have recently shown promising advancements, delivering state-of-the-art performance with rather lightweight architectures. In the context of graph-structured data, leveraging the eigenvectors of the graph Laplacian matrix for positional encoding is effective. Yet, the approach does not specify how to handle scenarios where edges in the input graph are missing. To this end, we propose a novel positional encoding technique, PerturbPE, that extracts consistent and regular components from the eigenbasis. Our method involves applying multiple perturbations and taking their average to extract the consistent and regular component from the eigenbasis. PerturbPE leverages the Rayleigh-Schrodinger Perturbation Theorem (RSPT) for calculating the perturbed eigenvectors. Employing this labeling technique enhances the robustness and generalizability of the model. Our results support our theoretical findings, e.g. our experimental analysis observed a performance enhancement of up to $12\%$ on the Human3.6M dataset in instances where occlusion resulted in the absence of one edge. Furthermore, our novel approach significantly enhances performance in scenarios where two edges are missing, setting a new benchmark for state-of-the-art. 
\end{abstract}

\section{Introduction}

Estimating the 3D pose of the human skeleton is crucial for understanding human motion and behavior, which facilitates high-level computer vision tasks such as action recognition~\cite{luvizon20182d} and augmented and virtual reality~\cite{han2018viton}. Nevertheless, estimating 3D human joint positions presents considerable obstacles. First, there is a scarcity of labeled datasets, as acquiring 3D annotations is costly. Additionally, challenges such as self-occlusions, complex joint inter-dependencies, and small and barely visible joints further complicate the estimation process.


To address the challenge of estimating 3D human poses, several strategies have been explored, including leveraging multi-view setups~\cite{rhodin2018learning}, utilizing synthetic data~\cite{peng2018jointly}, or incorporating motion analysis~\cite{shere2021temporally}. However, these methods can be cost-prohibitive, with multi-view configurations being impractical for real-world applications and the analysis of temporal data requiring significant resources. A more resource-efficient approach is lifting 2D-to-3D skeletons. The 2D human skeleton is a graph-structured data and thus Graph Convolutional Networks (GCNs) achieve state-of-the-art performance for 2D-to-3D human pose estimation by reducing the number of parameters in the order of magnitude, e.g.~\cite{azizi20223d}.

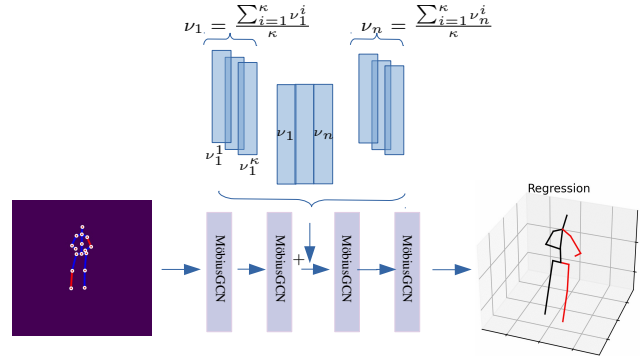
\begin{figure}
\begin{picture}(150,150)
\put(30,0){\includegraphics[scale=0.5]{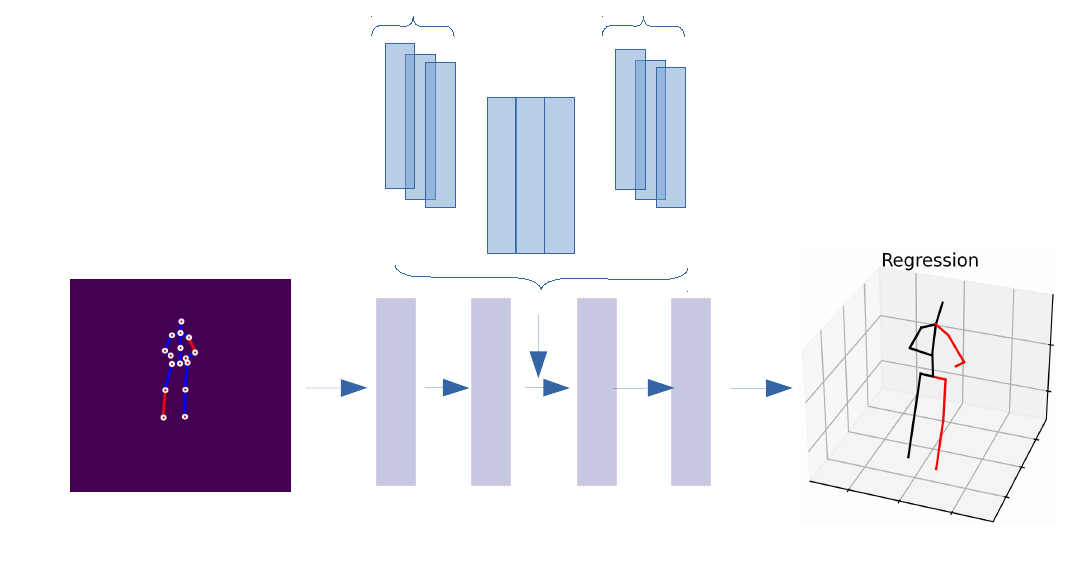}}
 \put(194.9, 54.8){\rotatebox{270}{\tiny{M\"{o}biusGCN}}}
 \put(172., 54.8){\rotatebox{270}{\tiny{M\"{o}biusGCN}}}
 \put(146.9, 54.8){\rotatebox{270}{\tiny{M\"{o}biusGCN}}}
 \put(124.2, 54.8){\rotatebox{270}{\tiny{M\"{o}biusGCN}}}
\put(112, 132){\scriptsize{$\nu_1 = \frac{\bigsum{i}{1}{\kappa}\nu_1^i}{\kappa}$}}
\put(179, 132){\scriptsize{$\nu_n = \frac{\bigsum{i}{1}{\kappa}\nu_n^i}{\kappa}$}}
\put(153, 44){\tiny{$+$}}
\put(147, 92){\tiny{$\nu_{1}$}}
\put(161, 92){\tiny{$\nu_{n}$}}
\put(120, 83.5){\tiny{$\nu_1^1$}}
\put(132.8, 79.5){\tiny{$\nu_1^{\kappa}$}}
\end{picture}
\caption{Given a human skeleton graph with only blue edges provided and red edges absent, we calculate the corresponding graph Laplacian matrix. By selectively removing several edges at random (the selected elements in the graph Laplacian matrix), we then compute the perturbed eigenvectors ($\nu_i$), which are subsequently averaged to identify consistent eigenbasis ($\frac{\bigsum{i}{1}{\kappa}\nu_i}{\kappa}$). Finally, we integrate the perturbed positional features into a graph neural network architecture. The perturbed eigenvectors are computed with Rayleigh-Schr\"{o}dinger Perturbation Theorem (RSPT).}
\end{figure}

\par Nevertheless, GCNs have limited expressivity. The seminal work~\cite{morris2019weisfeiler} shows that GCNs are as expressive as 1-Weisfeiler-Lehman (1-WL)~\cite{weisfeiler1968reduction}. Efforts aimed at augmenting the expressiveness of GCNs have pursued four main directions: aligning with the k-WL hierarchy~\cite{maron2018invariant}, enriching node features with identifiers, or exploiting structural information that cannot be captured by the WL test~\cite{bodnar2021weisfeiler}. Subgraph GNNs~\cite{bevilacqua2022equivariant} have emerged as a potential solution, enriching GNN features by encoding extracted subgraphs as novel features and incorporating them into the GNN architecture. One line of work addresses the limited expressivity of GCNs by positional encoding which utilizes the eigenbasis of graph Laplacian matrix~\cite{dwivedi2020generalization, lim2022sign}. However, if some edges are missing these methods are not applicable. This scenario may occur in human pose estimation when the subject is not completely visible due to obstructions, e.g., occlusion.
\par To address this problem, we propose a novel perturbed \textbf{P}ositional \textbf{E}ncoding (PerturbPE), where we label the nodes by considering that graph Laplacian matrix has regular and irregular parts~\cite{lu2015toward}.  
We extract the regular part of the graph Laplacian eigenbasis for the positional encoding, by applying multiple perturbations (, i.e., removing different sets of edges each time) to the initial graph Laplacian matrix. After computing the perturbed eigenvector each time, we average over the perturbed eigenvectors to extract the consistent part of the graph Laplacian eigenbasis with missed edges. Employing this labeling technique enhances the robustness and generalizability of the model.  
In other words, the eigenvectors can well reflect network structural features. As proved in~\cite{lu2015toward}, the regularity of the network is reflected in the consistency of structural features before and after a random removal of a small set of links. 

\par We summarize our main contributions as follows:
\begin{itemize}
    \item We provide a novel positional encoding, PerturbPE, to extract the regular part of the graph Laplacian eigenbasis for scenarios where some edges are missing in the input graph, e.g., scenarios where occlusion happens in the input human 2D skeleton.
    We utilize Rayleigh-Schr\"{o}dinger Perturbation Theorem (RSPT) to compute the eigenpairs. Its primary advantage is that it eliminates the need to calculate the entire eigenbasis.
    \item We attain state-of-the-art by while training only one neural network to adeptly manage all scenarios where specific edges (specific parts of the 2D human skeleton) are missing. 
    \item We also provide the state-of-the-art results of positional encoding for the task of 3D human pose estimation, achieving state-of-the-art 3D human pose estimation results, despite requiring the same number of model parameters as M\"{o}biusGCN~\cite{azizi20223d} on benchmark datasets.
\end{itemize}

\section{Related Work}
\label{sec:rel work}
\textbf{3D Human Pose Estimation}  
Classical methods for 3D human pose estimation primarily rely on had-engineered features and incorporate prior knowledge (~\emph{e.g.,}~\cite{h36m_pami, ramakrishna2012reconstructing, sminchisescu20083d}). While these techniques have achieved commendable outcomes, their main limitation is their inability to generalize.
Current state-of-the-art in computer vision, including 3D human pose estimation, primarily utilize Deep Neural Networks (DNNs)~\cite{li2019generating, luo2021multi, ma2021context}. These models operate under the assumption that input data exhibits characteristics of locality, stationarity, and multi-scalability. While DNNs excel in areas defined by Euclidean geometry, many challenges in the real world do not conform to this Euclidean framework. 
\par Graph Convolutional Networks (GCNs) have emerged as a powerful solution for these problem categories, encompassing two main types: spectral and spatial GCNs. Spectral GCNs operate on the principles of the Graph Fourier Transform to analyze graph signals via the graph's Laplacian matrix vector space. On the other hand, spatial GCNs focus on transforming features and aggregating neighborhood information directly on the graph,~\emph{e.g.,} Message Passing Neural Networks~\cite{gilmer2017neural} and GraphSAGE~\cite{hamilton2017inductive}.
\par  GCNs stand out for delivering high-quality results with a relatively small number of parameters in the task of 3D human pose estimation. Various studies~\cite{liu2020comprehensive, xu2021graph, zhaoCVPR19semantic}, have explored pose estimation using GCNs.~\citet{xu2021graph} introduced the Graph Stacked Hourglass Networks (GraphSH), a method that processes graph-structured features across multiple levels of human skeletal structures. Similarly,~\citet{liu2020comprehensive} delved into diverse strategies for feature transformation and the aggregation of neighborhood spatial features, highlighting the advantage of assigning unique weights to enhance a node's self-information.~\citet{zhaoCVPR19semantic} developed the semantic GCN (SemGCN) focusing on the concept of learning the adjacency matrix to capture semantic relationships between graph nodes. M\"{o}biusGCN~\cite{azizi20223d}, dramatically reduced parameter count to just $0.042\text{M}$ by leveraging the M\"{o}bius transformation to explicitly define joint transformations, showcasing an even more efficient model architecture for pose estimation.
\par \textbf{GCN Expressivity} Nonetheless, the expressiveness of GCNs is constrained, and to surmount this challenge, four principal approaches have been proposed. Evaluating the expressive capacity of GNNs requires dealing with the graph isomorphism, which has no P solution (NP-intermediate). 
 The 1-WL test~\cite{weisfeiler1968reduction}, effectively resolves graph isomorphism for a vast majority of graph-structured data. However, as indicated in research by~\citet{xu2018how, morris2019weisfeiler, li2022expressive}, the expressivity of MPNNs is limited to that of the 1-WL test. This limitation becomes particularly significant in real-world applications involving graph structures, as the 1-WL test cannot distinguish between certain graph features. Specifically, it fails in differentiating isomorphism in attributed regular graphs, measuring distances between nodes, and counting cycles within graphs~\cite{li2022expressive}.
Addressing these limitations, recent studies have provided four different approaches. The first approach involves adding random attributes to nodes with identical substructures. This method aims to provide uniqueness to these nodes, enhancing their capability to be differentiated. However, this advantage comes at the cost of reduced deterministic predictability, potentially leading to difficulties during inference~\cite{sato2021random}.
Furthermore, deterministic positional features (\emph{e.g.,}~\cite{zhang2018link}) argue that the incapability of the GNNs to encode the distance between nodes in the input graph raises the above issues and addresses them by injecting deterministic distance attributes. However, these methods assign different node features to isomorphic graphs and, thus, are not generalizable in inference time~\cite{li2022expressive}. The third strategy involves developing higher-order GNNs~\cite{maron2018invariant} to surpass the 1-WL test's expressivity limits. The fourth is adopting subgraph-based methods like ESAN~\cite{bevilacqua2022equivariant}, which enhance expressivity through selected subgraphs. 
\par Nevertheless, if some edges are missing, none of the previously mentioned methods can be applied to enhance expressivity. In the realm of 2D human pose estimation, this significant challenging scenario can happen when occlusion happens. When parts of the human body are occluded or hidden from view, accurately estimating the 3D human pose becomes more difficult. This problem is prevalent in real-world scenarios where objects may obstruct the view, or the camera angle limits visibility. 

\par To address this problem we propose a novel positional encoding. In the seminal work~\cite{dwivedi2020generalization}, it was proposed that incorporating graph Laplacian eigenvectors as positional features within the Graph Neural Network (GNN) architecture could improve its effectiveness, although this approach encountered obstacles such as sign ambiguity and eigenvalue multiplicity. To address these challenges, SignNet/BasisNet~\cite{lim2022sign} was developed. A Multilayer Perceptron (MLP) was employed to deal with the issue of multiplicity, while an approach using an even function—by amalgamating the function with its inverse—was implemented to tackle problems related to sign ambiguity. However, to the best of our knowledge, none of the previous works provided a solution in case some edges missing, which can be reduced to subgraph matching which is NP-complete. We propose extracting the consistent and robust part of the graph Laplacian matrix's eigenbasis by labeling the joints (graph nodes) utilizing the perturbed eigenbasis of the graph Laplacian matrix. To compute the perturbed eigenbasis, we employ the Rayleigh Schrodinger Perturbation Theory (RSPT). 

\textbf{Data Reduction}
Semi-supervised techniques are favored due to the tedious and costly nature of data annotation. M\"{o}biusGCN~\cite{azizi20223d} stands as the most efficient framework for 3D human pose estimation tasks to date, thereby diminishing the need for extensive annotated data for training. The advantage of light architectures is their ability to be trained with less data. Our novel perturbed positional encoding architecture, which does not add additional training parameters, maintains the framework's efficiency, enabling it to achieve leading results with a minimal amount of labeled data.

\section{Preliminaries}
To compute the consistent part of the graph Laplacian eigenbasis for the positional encoding, we use the perturbed eigenvectors. This process involves using the Rayleigh-Schrödinger Perturbation Theorem (RSPT) to determine the perturbed eigenvectors of the graph Laplacian matrix. Hence, we provide a concise overview of the RSPT. Moreover, while PerturbPE can enhance any GNN framework, our experiments are conducted using M\"{o}biusGCN, a model developed specifically for the task of 3D human pose estimation. Therefore, we also briefly overview the M\"{o}biusGCN.
 
\subsection{Rayleigh-Schr\"{o}dinger Perturbation Theory}
In mathematical terms, we are dealing with a discretized Laplacian-type operator represented by a real symmetric matrix that undergoes a minor symmetric linear perturbation. 
\begin{equation}
\mat{A}(\epsilon) = \mat{A}_0 + \epsilon \mat{A}_1. 
\label{aa}
\end{equation}
The Rayleigh-Schr\"{o}dinger Perturbation Theory (RSPT)~\cite{mccartin2009rayleigh} provides estimates for the eigenvalues and eigenvectors of the matrix $\mat{A}$ through a series of progressively higher-order adjustments to the eigenvalues and eigenvectors of the matrix $\mat{A_0}$. $\mat{A_0}$ is likewise real and symmetric but may possess multiple eigenvalues.

An advantage of RSPT is that it doesn't necessitate the full set of $\mat{A_0}$ and can be approached using the Moore-Penrose Pseudoinverse. The pseudoinverse need not be explicitly calculated since only pseudoinverse vector products are required. These may be efficiently be calculated by a combination of QR-factorization and Gaussian elimination. Since we are only concerned with real-symmetric matrices, the existence of a complete set of orthonormal eigenvectors is assured. 

\subsubsection{Reconstruction of Perturbed Matrix} To compute the eigenpairs of $\mat{A}$ possessing respective perturbation expansions 
\begin{equation}
\lambda_i(\epsilon) = \bigsum{k}{0}{\infty} \epsilon^k\lambda_i^{(k)} \quad
\vec{v}_i(\epsilon) = \bigsum{k}{0}{\infty} \epsilon^k\vec{v}_i^{(k)},
\label{eigenpairs}
\end{equation}
where $(i = 1, \dots, n)$ for sufficiently small $\epsilon$. Experimentally, we set both $\epsilon$ and $k$ to $1$.

Considering the eigenvalue problem and and taking into account the equations \ref{aa} and \ref{eigenpairs}, we have to solve the following recurrence relation 
\begin{equation}
(\mat{A}_0 - \lambda_i^{(0)}\mat{I})\vec{x}_i^{(k)} = - (\mat{A}_1 - \lambda_i^{(1)}\mat{I})\vec{x}_i^{(k-1)} + \bigsum{j}{0}{k-2} \lambda_i^{k-j} \vec{x}_i^{(j)}
\label{recursive}
\end{equation}
 for $ (k = 1, \dots, \infty; i = 1, \dots, n)$.
 
\par The solution is either degenerate or non-degenrate.

\subsubsection{Nondegenerate Case} By assuming all the eigenvalues are distinct the eigenpairs are computed as follows. 

If $j$ is odd, then
\begin{equation}
    \lambda_i^{2j+1} = \langle \vec{x}_i^{(j)}, \mat{A}_1 \vec{x}_i^{(j)} \rangle - \bigsum{\mu}{0}{j}\bigsum{\nu}{1}{j}\lambda_i^{(2j+1-\mu-\nu))} \langle \vec{x}_i^{(\nu)}, \vec{x}_i^{(\mu)} \rangle.
\end{equation}
If $j$ is even, then
\begin{equation}
    \lambda_i^{2j} = \langle \vec{x}_i^{(j-1)}, \mat{A}_1 \vec{x}_i^{(j)} \rangle - \bigsum{\mu}{0}{j}\bigsum{\nu}{1}{j}\lambda_i^{(2j-\mu-\nu))} \langle \vec{x}_i^{(\nu)}, \vec{x}_i^{(\mu)} \rangle.
\end{equation}
The corresponding eigenvector is computed as follows
\begin{equation}
    \vec{x}_i^{(k)} = (\mat{A}_0 - \lambda^{(0)}_i\mat{I})^{\dagger}[-(\mat{A}_1- \lambda_i^{(1)}\mat{I})\vec{x}_i^{(k-1)}+ \bigsum{j}{0}{k-2}\lambda_i^{(k-j)}\vec{x}_i^{(j)}]
\end{equation}
The unperturbed eigenvectors are assumed to have been normalized to unity so that $\lambda_i^{(0)} = \langle \vec{x}_i^{(0)}, \mat{A}_0\vec{x}_i^{0}  \rangle$. 

\subsubsection{Degenerate Case}
For the calculation of perturbed eigenpairs in scenarios involving multiplicity, $\lambda^{(0)}_1 = \lambda^{(0)}_2 = \dots = \lambda^{(0)}_m = \lambda^{(0)}$, accompanied by $m$ known orthonormal eigenvectors $\vec{v}^{(0)}_1,\dots, \vec{v}^{(0)}_m$ and the assumption of first-order degeneracy which ensures the uniqueness of the first-order eigenvalues, the calculation of these perturbed eigenvalues and their corresponding degenerate eigenvectors are achieved by determining appropriate linear combinations of 
   \begin{equation}
    \vec{y}^{(0)}_i = a^{(i)}_1\vec{v}^{(0)}_1+ a^{(i)}_2\vec{v}^{(0)}_2+ a^{(i)}_3\vec{v}^{(0)}_3+ \dots+ a^{(i)}_m\vec{v}^{(0)}_m, 
    \label{multiplicity}
    \end{equation}

To have a solution for Equation~\eqref{recursive} in this scenario, it is necessary and sufficient that for each fixed $i$, 
\begin{equation}
     \langle x^{(0)}_{\mu}, (\mat{A}_1 - \lambda_i^{(1)}\mat{I})\vec{y}_i^{(0)} \rangle = 0 \quad (\mu = 1, \dots, m)
\end{equation}
By replacing Equation~\eqref{multiplicity}, 
\[
 \begin{bmatrix}
\langle \vec{x}_1^{(0)}, \mat{A}_1\vec{x}_1^{(0)} \rangle & \dots & \langle \vec{x}_1^{(0)}, \mat{A}_1\vec{x}_m^{(0)} \rangle \\
\vdots & \ddots &\vdots\\
\langle \vec{x}_m^{(0)}, \mat{A}_1\vec{x}_1^{(0)} \rangle & \dots & \langle \vec{x}_m^{(0)}, \mat{A}_1\vec{x}_m^{(0)} \rangle
\end{bmatrix}
\begin{bmatrix}
    a_1^{(i)} \\
    \vdots \\
    a_m^{(i)}
\end{bmatrix} = \lambda_i^{(1)} \begin{bmatrix}
    a_1^{(i)} \\
    \vdots \\
    a_m^{(i)}
\end{bmatrix}
\]
the corresponding eigenvector of each $\lambda_i^{(1)}$ becomes $[ a_1^{(i)}, \dots, a_m^{(i)}]\transpose$.

\subsection{Spectral Graph Convolutional Network}
\subsubsection{Graph Definitions} 
Given a graph $\mathcal{G}(V, E)$ with vertices $V = \{\upsilon_1, \dots, \upsilon_N\}$ and edges $E = \{e_1, \dots, e_M\}$, where $e_j = (\upsilon_i, \upsilon_k)$ with $\upsilon_i, \upsilon_k \in V$. The adjacency matrix $\mathbf{A}$ marks $1$ for connected vertices and $0$ otherwise. The degree matrix $\mathbf{D}$ is diagonal, listing vertex degrees $\mathbf{D}{ii}$ for $\upsilon{i}$. Graph $\mathbf{A}$ is symmetric for undirected graphs. The graph Laplacian $\mat{L} = \mat{D} - \mat{A}$, and its normalized form $\Bar{\mat{L}} = \mat{I} - \mat{D}^{-\frac{1}{2}}\mat{A}\mat{D}^{-\frac{1}{2}}$, with $\mathbf{I}$ the identity matrix, are key in analyzing graph structure. $\Bar{\mathbf{L}}$ is symmetric and positive semi-definite with ordered, real, non-negative eigenvalues ${\lambda_i}$ and orthonormal eigenvectors ${\mathbf{u}_i}$. A graph signal $\mathbf{x} \in \mathbb{R}^N$ assigns values to vertices, and $\mathbf{X} \in \mathbb{R}^{N \times d}$ represents a $d$-dimensional signal on $\mathcal{G}$~\cite{shuman2013emerging}.

\iftrue
\subsubsection{Graph Fourier Transform} Graph signals $\textstyle\mathbf{x} \in \mathbb{R}^N$ admit a graph Fourier expansion $\mathbf{x}=\sum_{i=1}^N \langle \mathbf{u}_{i},\mathbf{x}\rangle \mathbf{u}_i$, where $\mathbf{u}_i, i=1,\dots,N$ are the eigenvectors of the graph Laplacian~\cite{shuman2013emerging}. Eigenvalues and eigenvectors of the graph Laplacian matrix are analogous to frequencies and sinusoidal basis functions in the classical Fourier series expansion. 

\subsubsection{Spectral Graph Convolutional Network} Spectral GCNs~\cite{bruna2013spectral} build upon the graph Fourier transform. Let $\vec{x}$ be the graph signal and $\vec{y}$ be the graph filter on graph $\mathcal{G}$. The graph convolution $\ast_\mathcal{G}$ can be defined as:
\begin{equation}
\vec{x} \ast_\mathcal{G} \vec{y} = \mat{U}\operatorname{diag}(\mat{U}\transpose\vec{y})\mat{U}\transpose\vec{x},
\end{equation}
where the matrix $\mat{U}$ contains the eigenvectors of the normalized graph Laplacian and $\odot$ is the Hadamard product. This can also be written as 
\begin{equation}
\vec{x} \;\ast_\mathcal{G} \; g_\theta = \mat{U}g_{\theta}(\mat{\Lambda})\mat{U}\transpose\vec{x},
\end{equation}
 where $g_{\theta}(\mat{\Lambda})$ is a diagonal matrix consisting of the learnable parameters, and is a function of the eigenvalues $\mat{\Lambda}$.
We utilize M\"{o}biusGCN~\cite{azizi20223d}, which defines the function $g_{\theta}$ to be M\"{o}bius transformation.

Thus a M\"obiusGCN block is
\begin{equation}
    \mat{Z} = \sigma(2\Re \{\mat{U} \mob(\mat{\Lambda})\mat{U}\transpose\mat{X}\mat{W}\} + \vec{b}),
\label{block}
\end{equation}
where $\mathbf{Z} \in \mathbb{R}^{N\times F}$ is the convolved signal matrix,
$\sigma$ is a nonlinearity  (\eg \text{ReLU}~\cite{nair2010rectified}), and $\vec{b}$ is a bias term and the graph signal matrix $\mathbf{X} \in \mathbb{C}^{N\times d}$ with $d$ input channels (\ie a $d$-dimensional feature vector for every node) and $\mathbf{W} \in \mathbb{C}^{d \times F}$ feature maps. 
\section{Method}
\subsubsection{Eigenvector Positional Encoding}
We aim to compute the consistent part of graph Laplacian eigenbasis to enhance generalizability, specifically in cases where the occlusions occur and the complete 2D human skeleton graph is not provided as an input to the architecture. To compute the positional encoding in such scenarios, we apply the Rayleigh-Schr\"{o}dinger Perturbation Theory (RSPT) multiple times ($\kappa$-times) independently. During each iteration, we randomly eliminate a number of edges from the graph's Laplacian matrix (denoted as $\mat{A}_0$ in Equation~\eqref{aa}), and then compute the perturbation with respect to it (we update  $\mat{A}_1$ with eliminated edges). By executing the RSPT for each of these iterations and averaging over them, 
 the regular part of the $\kappa$ eigenvectors can be extracted with 
\begin{equation}
 \vec{p} = \frac{\bigsum{i}{1}{\kappa}\vec{v}_i}{\kappa},
\end{equation}
where $\vec{v}_i$ is the $i^{th}$ perturbed eigenvector.

After executing the algorithm $\kappa$ times and averaging the outcomes to isolate the stable elements of the graph Laplacian matrix's eigenbasis, we employ the following positional encoding technique to incorporate the features into the architecture.

\subsubsection{Positional Features}
We incorporate positional features, similar to~\cite{dwivedi2020generalization}, into the architecture through the subsequent method.
\begin{flalign}
    \mat{X}^{\ell} &= \sigma(f(\mat{Z}^{\ell} + \mat{P}))
\end{flalign}
where $\mat{P} \in \mathbb{R}^{N \times N}$ is the PerturbPE positional encoding computed with the RSPT.
For each vector, we define the positional features, denoted as $\vec{p}$, as the mean of the perturbed eigenvectors where it contains the consistent regular part of the graph Laplacian eigenbasis. The function $f$ represents a Multilayer Perceptron (MLP) that is applied to both node and positional features. Therefore,~\autoref{block} becomes
\begin{equation}
    \mat{Z}^{\ell+1} = \sigma(2\Re \{\mat{U} \mob(\mat{\Lambda})\mat{U}\transpose\sigma(f(\mat{Z}^{\ell} + \mat{P}))\mat{W}^{\ell+1}\} + \vec{b}).
\label{block-p}
\end{equation}

\subsubsection{Masked Condition Strategy}
In our experiments, we operate under the assumption that each sample may lack certain components of the human skeleton, specifically that up to two edges between joints might be missing randomly during both testing and training phases. However, we assume that although some edges are missed the total number of joints is known.

\section{Experimental Results}
\subsection{Datasets and Evaluation Protocols}
We employ the widely recognized Human3.6M motion capture dataset for our study~\cite{h36m_pami}. This extensive dataset encompasses over 3.6 million images, collected from 11 participants engaging in 15 distinct activities, captured through four calibrated RGB cameras. This setup was meticulously designed to ensure a comprehensive capture of each subject's movements, both during the training and testing phases. In alignment with previous works~(\emph{e.g.,}~\cite{azizi20223d, martinez2017simple, Tekin_2017_ICCV, sun2017compositional, pavlakos2018ordinal, zhaoCVPR19semantic, sharma2019monocular, xu2021graph}), our experimental framework utilizes the data from five subjects (S1, S5, S6, S7, S8) for model training purposes, while reserving two subjects (S9 and S11) for the testing phase. Each sample is independently analyzed, reflecting the unique viewpoints provided by each camera. 

\par To evaluate our model's ability to generalize, we employ the MPI-INF-3DHP dataset~\cite{mono-3dhp2017}. This dataset features six subjects tested across three distinct settings: a studio with a green screen (GS), a studio lacking a green screen (noGS), and an outdoor environment (Outdoor). It's important to mention that for the experiments conducted using the MPI-INF-3DHP dataset, training was only performed on the Human3.6M dataset.
\par Following the methodology of prior research~\cite{martinez2017simple, Tekin_2017_ICCV, sun2017compositional, zhaoCVPR19semantic, xu2021graph}, we adopt the MPJPE protocol, Protocol~\#1. The MPJPE is the mean per joint position error in millimeters between predicted joint positions and ground truth joint positions after aligning the pre-defined root joints (~\emph{i.e.,} the pelvis joint). Note that some works~(\emph{e.g.,}~\cite{pavllo20193d, liu2020comprehensive}) use the P-MPJPE metric, which reports the error after a rigid transformation to align the predictions with the ground truth joints. However, we deliberately chose the standard MPJPE metric as it is more challenging and the more equitable basis it provides for comparing our work with previous research.

In assessing performance on the MPI-INF-3DHP test set, in alignment with prior research~\cite{xu2021graph, luo2018}, we adopt the 3D Percentage of Correct Keypoints (3D PCK) with a $150\text{mm}$ threshold~\cite{mono-3dhp2017}. This measure allows us to accurately gauge the accuracy of our 3D joint predictions within a specified error margin, offering a comprehensive view of our model's performance in comparison to the benchmarks set by preceding studies.


\subsection{Implementation Details}
\textbf{2D Pose Estimation.} 
PerturbPE receives 2D joint positions as inputs, which are independently estimated from the RGB images captured by all four cameras. PerturbPE operates independently from any off-the-shelf architecture employed for the estimation of 2D joint positions. Although CPN~\cite{chen2018cascaded} provides better 2D human skeleton estimation, similar to previous works~\cite{martinez2017simple, zhaoCVPR19semantic}, we use the stacked hourglass architecture~\cite{newell2016stacked} to estimate the 2D joint positions. The Hourglass architecture is a type of autoencoder architecture that incorporates multiple skip connections at various intervals. In line with~\cite{zhaoCVPR19semantic}, the stacked hourglass network undergoes initial pre-training on the MPII dataset~\cite{andriluka14cvpr}, followed by subsequent fine-tuning using the Human3.6M dataset~\cite{h36m_pami}. As detailed by~\citet{pavllo20193d}, the input joints are adjusted to fit within image coordinates and are normalized to the range of $[-1, 1]$.

\textbf{3D Pose Estimation.} The Human3.6M dataset~\cite{h36m_pami} provides ground truth 3D joint positions in world coordinates. To align with previous works~\cite{azizi20223d, zhaoCVPR19semantic}, we utilize camera calibration parameters to transform these joint positions into camera space. Furtheremore, when training the pipeline, akin to previous studies~\cite{azizi20223d, martinez2017simple}a predefined joint (the pelvis joint) as the center of the coordinate system is selected. 

We trained PerturbPE using Adam optimizer~\cite{kingma2014adam} with an initial learning rate of $0.001$ and mini-batches of size $64$. Our neural network pipeline, which operates in the complex-valued domain, is built upon the PyTorch framework~\cite{NEURIPS2019_9015}, which leverages Wirtinger calculus~\cite{kreutz2009complex} to enable backpropagation within the complex-valued domain.

We adopt M\"{o}biusGCN as our baseline architecture due to its lightweight nature and impressive accuracy. In our experiments, we utilize eight M\"{o}biusGCN blocks. Each block, excluding the first and last blocks with input and output channels set to $2$ and $3$ respectively, consists of either $128$ channels (yielding $0.16$ million parameters) or $192$ channels (resulting in $0.66$ million parameters). Furthermore, we incorporate positional encoding features in the BLA block, followed by a subsequent linear layer.

Same as~\cite{martinez2017simple, zhaoCVPR19semantic}, we predict the normalized locations~\cite{martinez2017simple, zhaoCVPR19semantic, pavlakos2017coarse, liu2020comprehensive} of $16$ joints (~\emph{i.e.,} without the 'Neck/Nose' joint) in 3D and use the mean squared error (MSE) loss between the 3D ground truth joint locations $\mathcal{Y}$ and our predictions $\mathcal{\hat{Y}}$
\begin{equation}
    \mathcal{L}(\mathcal{Y}, \mathcal{\hat{Y}}) = \sum_{i=1}^{k}(\mathcal{Y}_i-\mathcal{\hat{Y}}_i)^2,
\end{equation}
where $k$ is the number of joints~\cite{martinez2017simple, pavllo20193d}. Furthermore, similar to~\cite{azizi20223d, poier2018learning}, to let the architecture differentiate between different 3D poses with the same 2D pose, the center of mass of the subject is provided as an additional input. Please note that during inference the scale of the outputs is calibrated by forcing the sum of the length of all 3D bones to be equal to a canonical skeleton~\cite{azizi20223d, zhou2017towards, zhou2018monocap}. 

All of our experiments were conducted using a PyTorch framework~\cite{NEURIPS2019_9015} on an NVIDIA GeForce RTX 2080 GPU.
\subsubsection{RSPT Perturbed Eigenvectors} In our experiments we consider computing the perturbed eigenvectors with different scenarios. We consider scenarios where zero, one, or two random edges are removed for computing the perturbed eigenvectors from the input occluded 2D human skeleton graph. Specifically, in our experiments, we consider graph Laplacian eigenvectors~\cite{dwivedi2020generalization} positional encoding, RSPT eigenbasis with zero edge is missed for perturbation, one edge is missed for perturbation, or two edges are missed for computing the perturbation with RSPT.

\begin{table*}
\centering
\resizebox{12.5CM}{!}{%
 \begin{tabular}{l| c c c c c c c c c c c c c c c c c} 
 \hline
 \textbf{Protocol~\#1} &\text{\# Param.} & \text{Dir.} & \text{Disc.} & \text{Eat} & \text{Greet} & \text{Phone} & \text{Photo} & \text{Pose} & \text{Purch.} & \text{Sit} & \text{SitD.} & \text{Smoke} & \text{Wait} & \text{WalkD.} & \text{Walk} & \text{WalkT.} & \textbf{Average} \\ [0.5ex] 
 \hline
  ~\citet{martinez2017simple}\hypersetup{citecolor=black}   & 4M & 51.8 & 56.2 & 58.1 & 59.0 & 69.5 & 78.4 & 55.2 & 58.1 & 74.0 & \phantom{0}94.6 & 62.3 & 59.1 & 65.1 & 49.5 & 52.4 & 62.9  \\ 
  ~\citet{Tekin_2017_ICCV}\hypersetup{citecolor=black}  & n/a &54.2 &61.4 &60.2 &61.2& 79.4 &78.3& 63.1& 81.6& 70.1& 107.3& 69.3 &70.3 &74.3 &51.8& 63.2& 69.7\\
  ~\citet{sun2017compositional} \hypersetup{citecolor=black}& n/a& 52.8& 54.8& 54.2& 54.3 &61.8& 67.2& 53.1 &53.6& 71.7 &\phantom{0}86.7& 61.5& 53.4& 61.6& 47.1& 53.4& 59.1\\
 ~\citet{yang20183d}\hypersetup{citecolor=black}  & n/a&
  51.5 &58.9 &50.4& 57.0 &62.1 &65.4& 49.8 &52.7 &69.2 &\phantom{0}85.2 &57.4& 58.4 &\textbf{43.6}& 60.1& 47.7& 58.6\\
 ~\citet{hossain2018exploiting}\hypersetup{citecolor=black} & 16.96M & 48.4 & 50.7 & 57.2 & 55.2 & 63.1 & 72.6 & 53.0 & 51.7 & 66.1 & \phantom{0}80.9 & 59.0 & 57.3 & 62.4 & 46.6 & 49.6 & 58.3  \\
 ~\citet{fang2018learning}\hypersetup{citecolor=black}  & n/a &  50.1& 54.3& 57.0 &57.1 &66.6& 73.3 &53.4& 55.7& 72.8& \phantom{0}88.6& 60.3& 57.7& 62.7 &47.5& 50.6& 60.4\\
 ~\citet{pavlakos2018ordinal}\hypersetup{citecolor=black}  & n/a & 48.5 & 54.4 & 54.5 & 52.0 & 59.4 & 65.3 & 49.9 & 52.9 &  65.8 & 71.1 & 56.6 & 52.9 & 60.9 & 44.7 & 47.8 & 56.2  \\
 SemGCN~\citet{zhaoCVPR19semantic}\hypersetup{citecolor=black} & 0.43M &  48.2& 60.8& 51.8& 64.0 &64.6& \textbf{53.6}& 51.1& 67.4& 88.7& \phantom{0}\textbf{57.7}& 73.2& 65.6& 48.9& 64.8& 51.9& 60.8\\
 ~\citet{sharma2019monocular}\hypersetup{citecolor=black}  & n/a & 48.6 & 54.5 & 54.2 & 55.7 & 62.2 & 72.0 & 50.5 & 54.3 & 70.0 & \phantom{0}78.3 & 58.1 & 55.4 & 61.4 & 45.2 & 49.7 & 58.0  \\ 
  GraphSH~\cite{xu2021graph}\hypersetup{citecolor=black} $\ast$ & 3.7M & \textbf{45.2} & \textbf{49.9} & 47.5 & 50.9 & \underline{54.9} & 66.1 & 48.5 & 46.3 & \underline{59.7} &\phantom{0}71.5 & \underline{51.4} & \underline{48.6} & 53.9 & \underline{39.9} & \underline{44.1} & \underline{51.9} \\ 
 M\"{o}biusGCN~\cite{azizi20223d}~(HG) & \textbf{0.16M} & 46.7 & 60.7 & \underline{47.3} &\underline{50.7} &64.1  & 61.5 &\underline{46.2}  &\underline{45.3}  &67.1 &\phantom{0}80.4 & 54.6 & 51.4 & 55.4 & 43.2 & 48.6 & 52.1  \\
 Ours~(HG) & \underline{0.66M} & \underline{45.9} & \underline{50.1}& \textbf{41.2} & \textbf{43.2}& \textbf{52.7} & \underline{57.4} &\textbf{43.0} & \underline{38.4} &\textbf{55.4} & \underline{61.8}&\textbf{45.8}  &\textbf{46.8} & \underline{48.5} & \textbf{38.9} & \textbf{42.8} &  \textbf{50.8} \\

 \hline
   \citet{liu2020comprehensive}~(GT) & 4.2M & 36.8 & 40.3 & 33.0 & 36.3 & 37.5 & 45.0 & 39.7 & 34.9 & 40.3 & \phantom{0}47.7 & 37.4 & 38.5 & 38.6 & 29.6 & \underline{32.0} & 37.8  \\ 
  GraphSH~\cite{xu2021graph}~(GT) & 3.7M & 35.8 & \underline{38.1} & \underline{31.0} & 35.3 & \textbf{35.8} & \underline{43.2} & 37.3 & 31.7 & \underline{38.4} & \phantom{0}\underline{45.5} & \underline{35.4} & 36.7 & \underline{36.8} & \underline{27.9} & \textbf{30.7} & \underline{35.8}  \\
    SemGCN~\cite{zhaoCVPR19semantic}~(GT) & 0.43M & 37.8 & 49.4 & 37.6 & 40.9 & 45.1 & 41.4 & 40.1 & 48.3 & 50.1 & \phantom{0}42.2 & 53.5 & 44.3 & 40.5 & 47.3 & 39.0 & 43.8  \\
 M\"{o}biusGCN~\cite{azizi20223d}~(GT) & \textbf{0.16M} & \underline{31.2} & 46.9 & 32.5 & \underline{31.7} & 41.4 & 44.9 & \underline{33.9} & \underline{30.9}& 49.2 & \phantom{0}55.7 & 35.9 &\underline{36.1} & 37.5 &  29.07 & 33.1& 36.2  \\ 

\midrule

Ours~(GT) & \underline{0.66M} & \textbf{30.1}& \textbf{35.3}& \textbf{30.6} &\textbf{27.6 }& \underline{36.2} & \textbf{38.4 }&\textbf{30.7} & \textbf{30.3} &\textbf{35.9} &\textbf{40.7} &\textbf{32.9} & \textbf{34.9} &\textbf{35.2}  &\textbf{27.2} & \underline{32.0}  &  \textbf{32.7}\\ 
\hline
 \end{tabular}
}
\vspace{0.2cm}
\caption{
Quantitative Evaluation Using MPJPE (mm) on the Human3.6M~\cite{h36m_pami} Dataset under Protocol \#1, Highlighting Leading Performances. In the upper section, methods utilize stacked hourglass (HG) 2D estimates \cite{newell2016stacked}, with the exception of one approach using CPN~\cite{Chen_2018_CVPR} (denoted by $\ast$). The lower section compares methods based on 2D ground truth (GT) inputs. Best results are highlighted in \textbf{bold}, and the second-best are \underline{underlined}. Lower is better.
}
 \label{protocol1}
\end{table*}

\subsection{Complete 2D Human Skeleton}
In this study, we present a comparative analysis of PerturbPE's performance using a complete 2D human skeleton against the former leading techniques in 3D human pose estimation on the Human3.6M and MPI-INF-3DHP datasets. Our comparison utilizes two types of inputs: a) 2D poses estimated through the stacked hourglass architecture (HG)~\cite{newell2016stacked} and b) the 2D ground truth (GT).

\textbf{Comparison on Human3.6M.} ~\autoref{protocol1} presents a comparison between our PerturbPE method and the leading techniques as per Protocol \#1 in the Human3.6M dataset. Utilizing the eigenvector of the graph Laplacian matrix for positional encoding leads to enhanced performance, reducing the error rate from $34.1 \text{mm}$ to $33.4 \text{mm}$ without an increase in model complexity. By introducing perturbed eigenvectors to tackle the multiplicity issue, we improved further, lowering the error rate to $32.7\text{mm}$. Notably, these advancements in positional encoding are achieved without the addition of extra parameters, ensuring the model remains efficient while benefiting from these enhancements. To evaluate the efficacy of our PerturbPE, we conducted experiments using SemGCN~\cite{zhaoCVPR19semantic}. Please refer to the supplementary material in \autoref{semgcn-PerturbPE} for more details. 

\textbf{Comparison on MPI-INF-3DHP.} To assess the adaptability and robustness of our method, we conducted evaluations using the MPI-INF-3DHP dataset, despite our model, PerturbPE, being exclusively trained on the Human3.6M dataset. This choice allowed us to test the generalizability of our approach beyond the conditions for which it was directly trained. In a comparative analysis with M\"{o}biusGCN, PerturbPE exhibited outstanding performance, marking a notable improvement in the key metric of evaluation, PCK. With an identical configuration in terms of the number of parameters between the two models, our method demonstrated an overall enhancement in the PCK metric across various testing scenarios, improving the initial score of $80.0$ to an improved score of $82.0$. The significance of this improvement was even more pronounced in the most challenging conditions presented by outdoor scenarios, where our model achieved a state-of-the-art PCK score of $84.0$. This result underscores the efficacy of PerturbPE in handling complex real-world situations. 
The results are in~\autoref{mpi-inf-3dhp}.
\begin{center}
\resizebox{6.85CM}{!}{%
\begin{tabular}{l| c c c c c}
\toprule
Method & \# Parameters & GS & noGS & Outdoor & All(PCK) \\
\midrule
\citet{martinez2017simple} & 4.2M & 49.8 & 42.5 & 31.2 & 42.5 \\
\citet{mono-3dhp2017} & n/a & 70.8 & 62.3 & 58.8 & 64.7 \\
\citet{luo2018} & n/a & 71.3& 59.4& 65.7& 65.6 \\
\citet{yang20183d} & n/a & - & - & - & 69.0 \\
\citet{zhou2017towards} & n/a & 71.1 & 64.7 & 72.7 & 69.2 \\
\citet{ci2019optimizing} & n/a & 74.8& 70.8& 77.3 &74.0 \\
\citet{zhou2019hemlets} & n/a & 75.6& 71.3& 80.3 &75.3 \\
GraphSH~\cite{xu2021graph} & 3.7M & \textbf{81.5} & \textbf{81.7} & 75.2 & \underline{80.1} \\
M\"{o}biusGCN~\cite{azizi20223d} & \textbf{0.16M} & 79.2 & 77.3 & \underline{83.1} & 80.0 \\
\midrule
Ours & \textbf{0.16M}  &\underline{80.0} & \underline{79.0} & \textbf{84.0} & \textbf{82.0} \\
\bottomrule
\end{tabular}
}
\captionof{table}{Results on the MPI-INF-3DHP test set~\cite{mono-3dhp2017}. Best in bold, second-best underlined. All methods use 2D ground truth as input. Lower is better.}
\label{mpi-inf-3dhp}
 \end{center}
\textbf{Comparison to Previous GCNs.} ~\autoref{light-archs-gcn-comparisons} 
showcases how our approach stands up against previously established GCN models. By augmenting the channel count in each block of the M\"{o}biusGCN architecture from $128$ to $192$, we observed an enhancement in the MPJPE metric by $2.1\text{mm}$. Building on this improvement, our novel PerturbPE technique further advances MPJPE performance by an additional $1.4\text{mm}$, all while maintaining an equivalent number of parameters to M\"{o}biusGCN. This strategic enhancement effectively reduces the MPJPE from $34.1\text{mm}$ down to $32.7\text{mm}$, illustrating the efficacy of our modifications in refining pose estimation accuracy.
\begin{center}
\resizebox{5CM}{!}{%
\begin{tabular}{l|c c}
\toprule
 Method & \# Parameters & MPJPE  \\
\midrule
\citet{liu2020comprehensive}\hypersetup{citecolor=black} & 4.20M & 37.8 \\
GraphSH~\cite{xu2021graph}\hypersetup{citecolor=black} & 3.70M & 35.8 \\
\citet{liu2020comprehensive}\hypersetup{citecolor=black} & 1.05M & 40.1 \\
GraphSH~\cite{xu2021graph}\hypersetup{citecolor=black} & 0.44M & 39.2 \\
SemGCN~\cite{zhaoCVPR19semantic}\hypersetup{citecolor=black} & 0.43M & 43.8 \\
\citet{yan2018spatial}\hypersetup{citecolor=black} & 0.27M & 57.4 \\
\citet{velivckovic2017graph}\hypersetup{citecolor=black} & \textbf{0.16M} & 82.9 \\
M\"{o}biusGCN~\cite{azizi20223d}\hypersetup{citecolor=black}  & \textbf{0.16M} & 36.2 \\
M\"{o}biusGCN~\cite{azizi20223d}\hypersetup{citecolor=black}  & 0.66M & \underline{34.1}\\
\midrule


Ours  & 0.66M & \textbf{32.7}\\
\bottomrule
\end{tabular}
}
 \captionof{table}{Supervised quantitative comparison between GCN architectures on Human3.6M~\cite{h36m_pami} under Protocol~\#1. Best in bold, second-best underlined. All methods use 2D ground truth as input. Lower is better.}
\label{light-archs-gcn-comparisons}
 \end{center}
\textbf{PerturbPE with Reduced Dataset.} PerturbPE adeptly mirrors the parameter efficiency of the lightweight M\"{o}biusGCN model, reaping the advantages of a compact design that necessitates a smaller volume of training data. By incorporating our novel positional encoding technique, PerturbPE dramatically refines its performance from $44.7\text{mm}$ to $42.9\text{mm}$ with MPJPE metric, achieved by analyzing data from just three subjects. This is particularly advantageous given the significant expense involved in acquiring 3D ground truth annotations. Moreover, we demonstrate that by further reducing the subject count to two and then to one, the results still show an improvement from $50.9\text{mm}$ to $48.9\text{mm}$ and from $67.4\text{mm}$ to $66.4\text{mm}$, respectively. Results are detailed in~\autoref{tb-data-red}. All experiments were conducted using ground truth 2D human skeleton data as the input.

\begin{center}
\resizebox{5CM}{!}{%
 \begin{tabular}{ l| c| c |c} 
\toprule
          Subject & \# Parameters & M\"{o}biusGCN~\cite{azizi20223d} & PerturbPE\\ [0.5ex] 
 \midrule
 S1 &  0.15M & 44.7 & \textbf{42.9}\\
 S1 S5 & 0.15M &50.9 & \textbf{48.9}\\
 S1 S5 S6 & 0.15M &67.4& \textbf{66.4}\\
\midrule
 \end{tabular}
}
 \captionof{table}{Evaluating the effects of using fewer training subjects on Human3.6M~\cite{h36m_pami} under Protocol~\#1 (given 2D GT inputs). Lower is better.}
\label{tb-data-red}
 \end{center}
\subsection{Recover 3D pose from partial 2D observation}
 This section explores the impact of applying positional encoding through PerturbPE in scenarios where different numbers of edges are missing~(\emph{i.e.,} the problem reduced to subgraph matching). Initially, we examine scenarios with the absence of a single edge. Subsequently, we delve into the more challenging scenarios involving the omission of two edges.
 In the real world, these scenarios frequently arise since incomplete 2D pose estimates frequently occur in 2D human pose estimation. This typically happens when body parts are outside of the camera view or obscured by objects within the scene.


\subsubsection{2D Human Skeleton with One Random Edge Missing}

In this first experiment, we eliminate one random edge from the input 2D human skeleton. We experiment the PerturbPE positional encoding under three conditions: no edge, one edge, and two edges missing for computing the perturbed eigenvectors.

Initially, when we compute the perturbed eigenvectors with no edge missing, which leads to addressing multiplicity, the results enhance accuracy from $55.0\text{mm}$ to $51.4\text{mm}$. Further improvements achieved by averaging results from two applications of one-time perturbations, reducing the error to $49.0\text{mm}$. Doubling the perturbation with two missing edges further refined accuracy to $48.0$ mm, validating our theoretical approach.

This experiment demonstrates that increasing the frequency of perturbations and removing a greater number of edges leads to the extraction of the most robust components of the graph Laplacian eigenbasis. Consequently, leading to more consistent labeling, which in turn improves the outcomes during inference. However this improvement comes with the computational expense.

The results are demonstrated in~\autoref{1-edge-missed}, showcasing the effect of PerturbPE positional encoding in partially observed 2D human skeleton input graphs when one edge is missing, also compared to eigenvector labeling~\cite{dwivedi2020generalization}. 

\begin{center}
\resizebox{6.7CM}{!}{%
 \begin{tabular}{l|c} 
\toprule
 Method & MPJPE\\ [0.5ex] 
 \midrule
  PerturbPE Label w. Eigenvector  &  55.0   \\
  PerturbPE Label w. Perturbed Eigenvector w. multiplicity    &  51.4 \\
  PerturbPE Label w. Perturbed Eigenvector w. 1-edge perturb  &  49.0 \\
  PerturbPE Label w. Perturbed Eigenvector w. 2-edge perturb  &  \textbf{48.0} \\ 
 \bottomrule
 \end{tabular}
}
 \captionof{table}{Evaluating the effects of positional encoding with one edge missing on Human3.6M~\cite{h36m_pami} under Protocol~\#1 (given 2D GT inputs). Lower is better.}
\label{1-edge-missed}
 \end{center}
\begin{figure}
\centering
\subfloat{\includegraphics[width=0.3\textwidth]{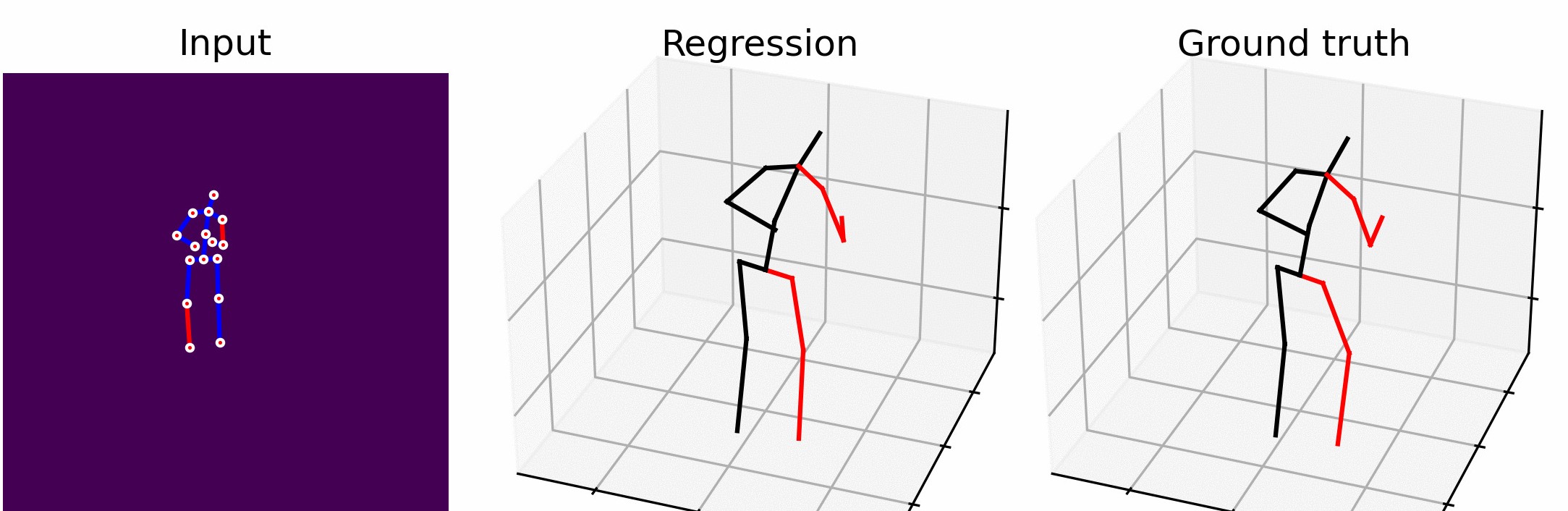}}
\quad
\subfloat{\includegraphics[width=0.3\textwidth]{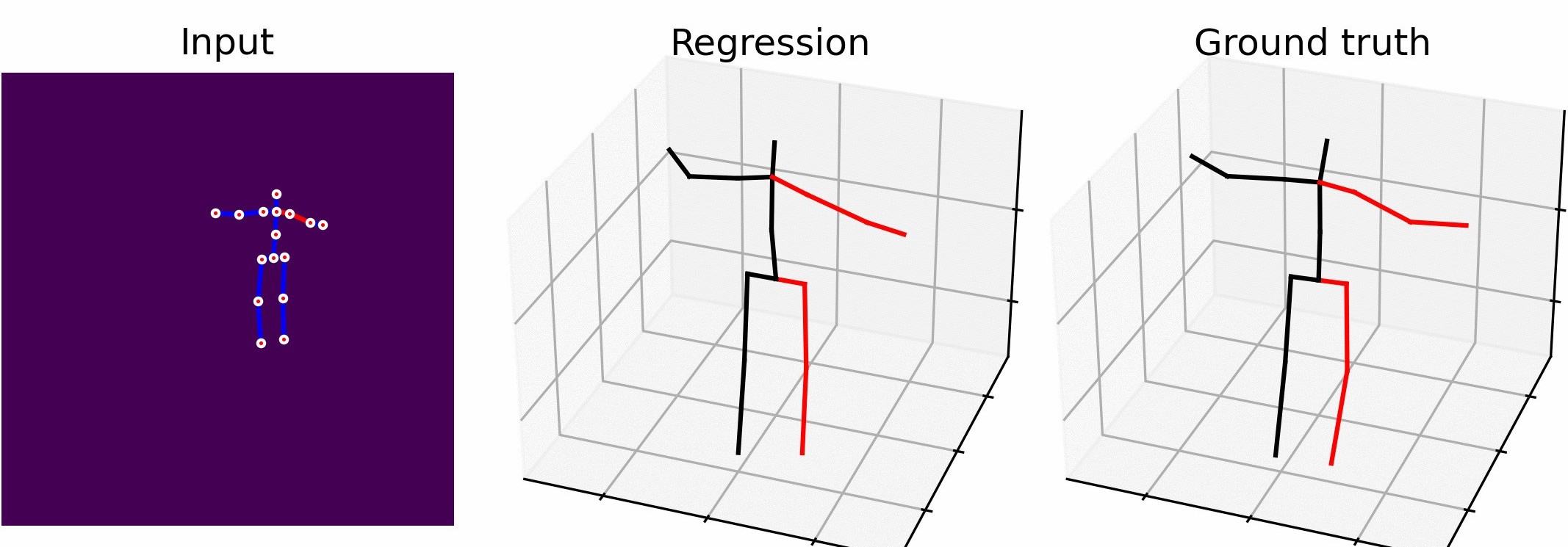}}
\quad
\subfloat{\includegraphics[width=0.3\textwidth]{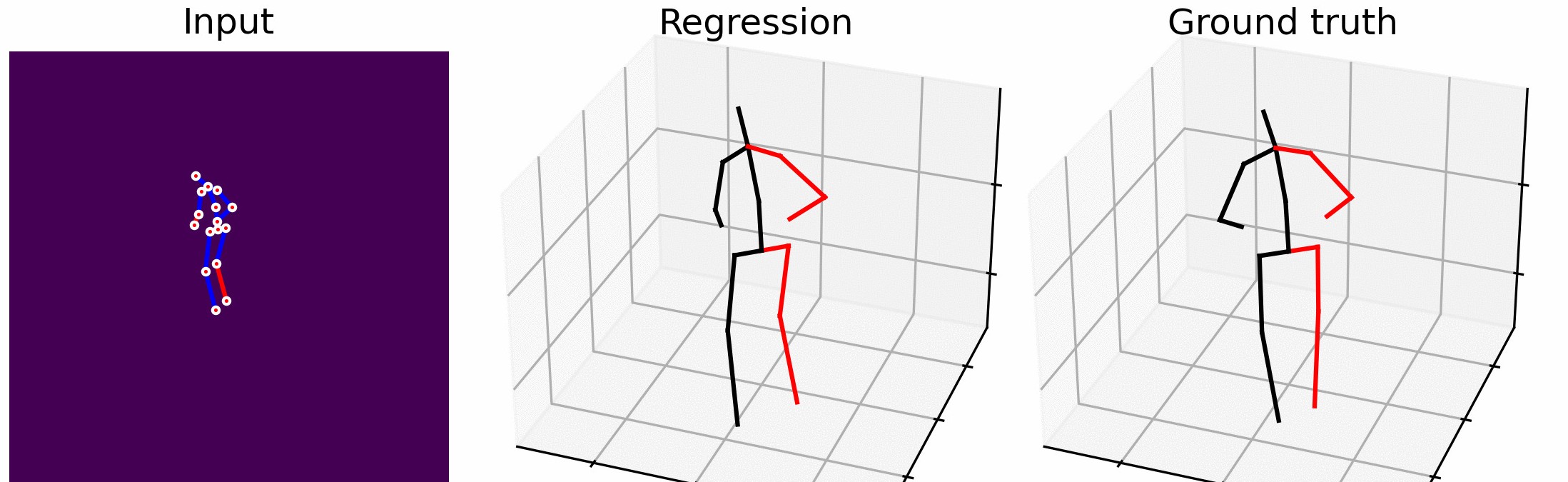}}\\ 
\subfloat{\includegraphics[width=0.3\textwidth]{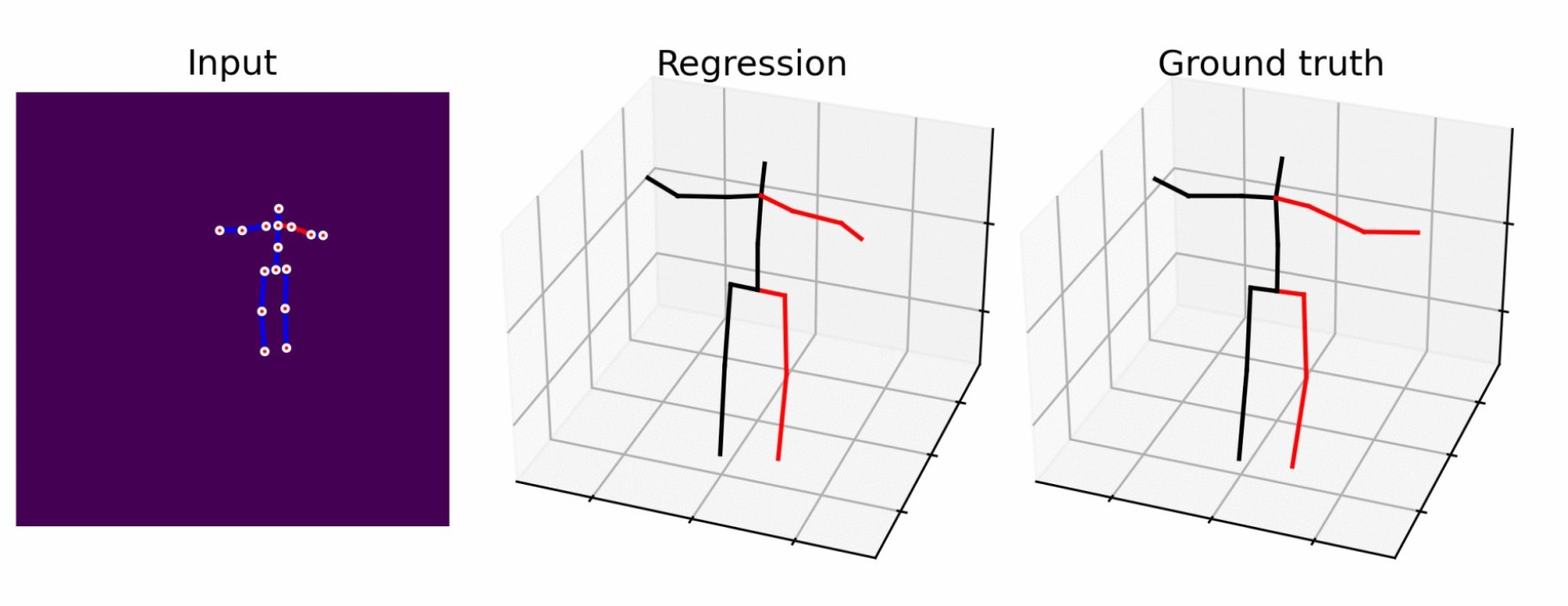}}
\quad
\subfloat{\includegraphics[width=0.3\textwidth]{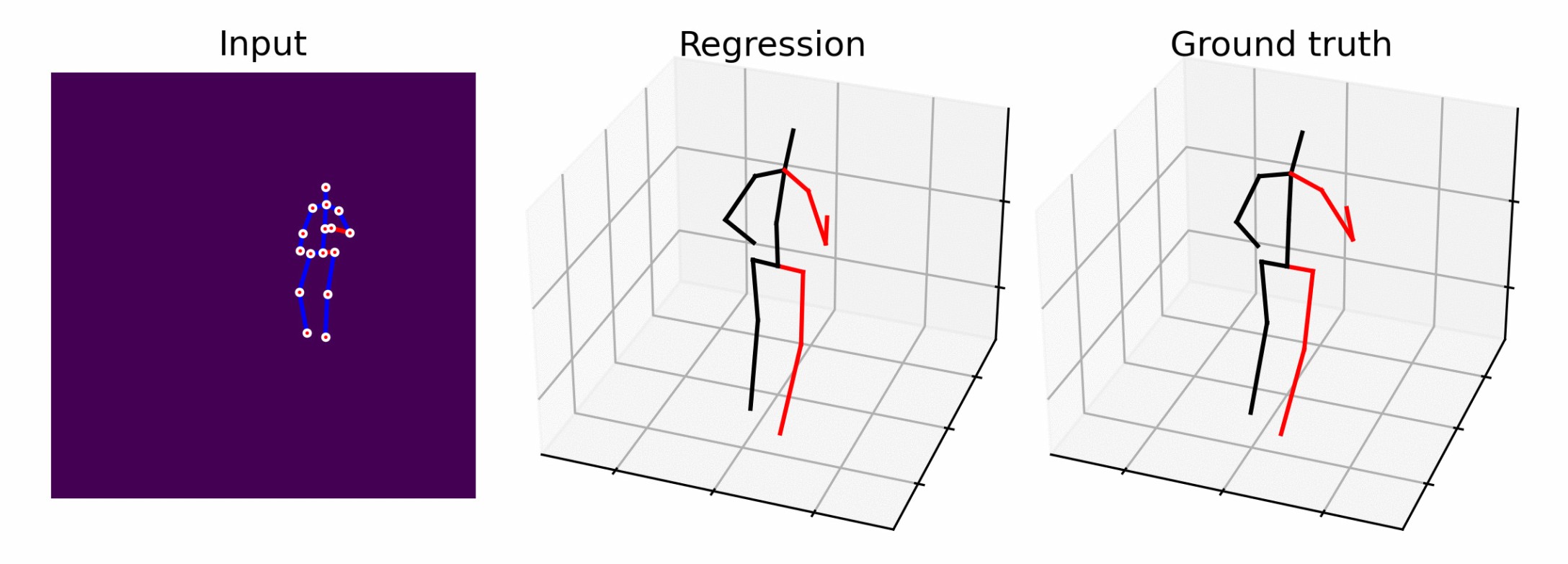}}
\quad
\subfloat{\includegraphics[width=0.3\textwidth]{img/output-0040.jpg}} \\
\caption{Qualitative self-occlusion results of PerturbPE on Human3.6M~\cite{h36m_pami}. This figure illustrates the PerturbPE's performance when trained to cope with the absence of any two arbitrary edges in the 2D human skeleton input. The network is trained for any two random edges missing in the input 2D human skeleton. A human skeleton graph with only blue edges provided and red edges absent.}
\label{qual-res-mpi-inf}
\end{figure}
\vspace{-30px}
\subsubsection{2D Human Skeleton with Two Random Edges Missing}
 In this experiment, we increase the difficulty of the task by removing two edges from the input 2D human skeleton. We experiment the PerturbPE positional encoding mainly under one condition: no edge is missing for computing the perturbed eigenvectors. In this scenario, we observe that when assigning labels by addressing multiplicity issues—the performance notably improves by approximately $10\%$ (decreasing the MPJPE from $60.00\text{mm}$ to $54.00\text{mm}$). The outcomes of this study are depicted in~\autoref{2-edge-missed}.

\par Further, we compare our results with GFPose~\cite{Ci_2023_CVPR}. While GFPose adopts a strategy of training distinct networks tailored to specific instances of missing body parts, our approach differs by training a single network that can handle any combination of missing edges. Despite this, our novel PerturbPE architecture significantly surpasses GFPose's performance, demonstrating the superior efficacy of our approach. Quantitative and qualitative results are shown in ~\autoref{2-edge-missed-comparison} and~\autoref{qual-res-mpi-inf}, respectively. In experiments with a 2D human skeleton missing arms, accuracy improved from $60.0\text{mm}$ to $58.6\text{mm}$. The absence of both legs further demonstrated the effectiveness of our PerturbPE method, enhancing performance to $52.4\text{mm}$. PerturbPE outperforms GFPose when either the right or left leg and arm are missing and shows better or similar results when any two edges are absent. 

\iftrue
\begin{center}
\resizebox{5.9CM}{!}{%
 \begin{tabular}{l|c} 
\toprule
 Method & MPJPE\\ [0.5ex] 
 \midrule
 M\"{o}biusGCN                                            &  \phantom{0}60.0 \\
 PerturbPE w. Perturbed Eigenvector w. 0-edge missed       &  \phantom{0}54.0 \\
 \bottomrule
 \end{tabular}
}
 \captionof{table}{Assessment of PerturbPE positional encoding impact on Human3.6M~\cite{h36m_pami} dataset performance with two edges removed, utilizing Protocol~\#1 with given 2D ground truth inputs. Lower is better.}
\label{2-edge-missed}
 \end{center}
 \fi

\textbf{Time Complexity} The computational complexity of the RSPT algorithm is $\mathcal{O}(n^3)$. Nevertheless, this level of complexity becomes acceptable for our purposes by taking into account the number of nodes present in a human skeleton and limiting ourselves to one order of perturbation (M\"{o}biusGCN has an inference time of 0.009 seconds per sample, while our method takes 0.010 seconds, demonstrating similar performance).

\begin{center}
\resizebox{5.6CM}{!}{%
 \begin{tabular}{l|c|c} 
\toprule
 Occ. Body Parts & Ours & GFPose~\cite{Ci_2023_CVPR} \\ [0.5ex] 
 \midrule
   2 Legs                                         & \textbf{52.4}  & 53.5\\
   2 Arms                                         & \textbf{58.6}  & 60.0\\
   Left Leg + Left Arm                            & \textbf{48.8}  & 54.6 \\
   Right Leg + Right Arm                          &  \textbf{44.6} & 53.1 \\   
 \bottomrule
 PerturbPE(Any Two Edges Missed)                           &  \textbf{54.0} &  \\ 
 \bottomrule
 \end{tabular}
}
 \captionof{table}{  Recover 3D pose from partial 2D observation: We train one model for two random missing edges with a masking strategy on Human3.6M dataset~\cite{h36m_pami} under Protocol~\#1 (given 2D GT inputs). Lower is better.}
\label{2-edge-missed-comparison}
 \end{center}

\section{Conclusions}
In this paper, we introduced the PerturbPE technique, a novel positional encoding method that leverages the Rayleigh-Schrodinger Perturbation Theorem (RSPT) to compute perturbed eigenvectors. This technique enables the extraction of consistent and regular components from the eigenbasis in cases where the input graph has missing edges, thereby enhancing the model's robustness and generalizability.
Our empirical evidence strongly supports our theoretical claims. Notably, we witnessed an improved performance of up to $12\%$ on the Human3.6M dataset when occlusion led to the absence of an edge. The performance improvement was even more significant in scenarios where two edges were missing, setting a new state-of-the-art benchmark.
While the initial results are promising, potential future work could involve refining the PerturbPE technique, investigating other potential applications of the RSPT in graph-structured data analysis, or exploring the scalability of our proposed method for larger datasets. Ultimately, this paper presents a novel encoding approach that boosts the capabilities of Graph Convolutional Networks (GCNs) in handling missing edges in the input graph-structured data, marking a significant stride in the field of 3D human pose estimation.
\clearpage  

%
%

\input{supplementary}
\bibliographystyle{plainnat}
\bibliography{egbib}
\clearpage

\end{document}

%% file: supplementary.tex
\begin{appendix}
\counterwithin{figure}{section}
\counterwithin{table}{section}
\section{Supplementary Material}
\iftrue
Below we'll explore one particular example (missing two legs) thoroughly.

\textbf{Two Legs are Missed}
The corresponding graph Laplacian matrix in this scenario when two edges (specifically legs; the ones in red) are missed is
 \begin{center}
 $\begin{bmatrix}
 3& \orange{-1}& 0& 0& \bl{-1}&  0&  0& \orange{-1}& 0& 0&  0&  0&  0&  0&  0&  0\\
\orange{-1}&  2&-1& 0&  0&  0&  0&  0& 0& 0&  0&  0&  0&  0&  0&  0\\
 0& -1& 2&\n{-1}&  0&  0&  0&  0& 0& 0&  0&  0&  0&  0&  0&  0\\
 0&  0&\n{-1}& 1&  0&  0&  0&  0& 0& 0&  0&  0&  0&  0&  0&  0\\
\bl{-1}&  0& 0& 0&  2& -1&  0&  0& 0& 0&  0&  0&  0&  0&  0&  0\\
 0&  0& 0& 0& -1&  1&  0&  0& 0& 0&  0&  0&  0&  0&  0&  0\\
 0&  0& 0& 0&  0&  0&  0&  0& 0& 0&  0&  0&  0&  0&  0&  0\\
\orange{-1}&  0& 0& 0&  0&  0&  0&  2&-1& 0&  0&  0&  0&  0&  0&  0\\
 0&  0& 0& 0&  0&  0&  0& -1& 4&-1& -1&  0&  0& -1&  0&  0\\
 0&  0& 0& 0&  0&  0&  0&  0&-1& 1&  0&  0&  0&  0&  0&  0\\
 0&  0& 0& 0&  0&  0&  0&  0&-1& 0&  2& -1&  0&  0&  0&  0\\
 0&  0& 0& 0&  0&  0&  0&  0& 0& 0& -1&  2& -1&  0&  0&  0\\
 0&  0& 0& 0&  0&  0&  0&  0& 0& 0&  0& -1&  1&  0&  0&  0\\
 0&  0& 0& 0&  0&  0&  0&  0&-1& 0&  0&  0&  0&  1&  \n{-1}&  0\\
 0&  0& 0& 0&  0&  0&  0&  0& 0& 0&  0&  0&  0&  \n{-1}&  1& -1\\
 0&  0& 0& 0&  0&  0&  0&  0& 0& 0&  0&  0&  0&  0& -1&  1\\
\end{bmatrix}$
\end{center}

\subsubsection{Perturbed with Zero Edges}
To compute the perturbed eigenvectors when only zero edges are missed (which means addressing multiplicity)
\begin{itemize}
    \item \textbf{Eigenvalue} 
    \begin{itemize}
        \item $\begin{bmatrix}
    0. &    0.  &   0.  &   0.10 & 0.27  & 0.36 & 1. &    1. &    1.46 & 1.68&
 & 2.00   &  2.35  & 3.08 & 3.15 & 4.32 & 5.21
\end{bmatrix}$
    \end{itemize}
\end{itemize}

In this scenario computing the perturbed eigenvectors contributes to resolving the issue of repeated eigenvalues for $0$ and $1$,  thereby leading to an improvement. 

\subsubsection{Perturbed with One Edge}
In this scenario, we perturb the matrix with one randomly selected edge~(\emph{e.g.,} the blue one in the graph Laplacian matrix) $\kappa$-times and compute the average to extract the consistent part of the graph Laplacian eigenbasis. 

\subsubsection{Pertubed with Two Edges}
In this scenario, we consider perturbing the graph Laplacian matrix with two edges~(\emph{e.g.,} the orange ones in the graph Laplacian matrix) $\kappa$-times and compute the average to extract the consistent part of the graph Laplacian eigenbasis.
\fi

\subsubsection{PerturbPE Efficacy on other GNNs} Our proposed method, PerturbPE, is designed as a positional encoding approach for scenarios where some edges are missing. These features can be easily integrated into other GNN methods as well. We conducted experiments on SemGCN~\cite{zhaoCVPR19semantic} using PerturbPE, and the results are in~\autoref{semgcn-PerturbPE}. We made minor modifications to adapt and integrate PerturbPE to SemGCN. We reduced the features in the penultimate block to 16 and summed our positional features within this block.
\vspace{-0.05cm}
\begin{center}
\resizebox{4.9CM}{!}{%
 \begin{tabular}{l|c|c} 
\toprule
 & \# Params & MPJPE \\ [0.5ex] 
 \midrule
   SemGCN    &           $0.21\text{M}$                        & 43.1   \\
   SemGCN + PerturbPE  & $0.21\text{M}$                        & \textbf{42.0} \\
 
 \bottomrule
 \end{tabular}
}
 \captionof{table}{PerturbPE Effect on SemGCN~\cite{zhaoCVPR19semantic}}
\label{semgcn-PerturbPE}
 \end{center}

\subsubsection{Recover 3D pose from partial 3D observation}
Similar to GFPose~\cite{Ci_2023_CVPR}, we also consider the case where the partial 3D is observed. which is a common real-world scenario. Results are presented in~\autoref{3d-pose-estimate}. 

Similar to the findings in GFPose~\cite{Ci_2023_CVPR} the most challenging situation is observed when the left arm is missing. Nevertheless, under these circumstances, the introduction of perturbPE improves the accuracy, achieving a precision of $6.0\,\text{mm}$. In every other tested scenario, perturbPE consistently excels over GFPose.
\begin{table}
\centering
\resizebox{4.0CM}{!}{%
 \begin{tabular}{l|c|c} 
\toprule
 Occ. Body Parts & Ours & GFPose~\cite{Ci_2023_CVPR} \\ [0.5ex] 
 \midrule
   Right Leg                          & \textbf{4.9} & 5.2 \\
   Left Leg                           & \textbf{2.2}  & 5.8 \\
   Left Arm                           & \textbf{6.0}  & 9.4\\
   Right Arm                          & \textbf{4.3}  & 8.9 \\
 \bottomrule
 \end{tabular}
}
\vspace{0.5cm}
 \captionof{table}{  Recover 3D pose from partial 3D observation: We train one model on Human3.6M dataset~\cite{h36m_pami} under Protocol~\#1 (given 3D GT inputs). Lower is better.}
\label{3d-pose-estimate}
\end{table}
\end{appendix}